\documentclass[letterpaper, 10 pt, conference]{ieeeconf} 
\IEEEoverridecommandlockouts
\usepackage{moreverb,url}
\usepackage[colorlinks,bookmarksopen,bookmarksnumbered,citecolor=red,urlcolor=red]{hyperref}
\usepackage{siunitx}
\usepackage{url}
\usepackage{svg}
\usepackage{multicol}
\usepackage{soul}
\usepackage{color}
\usepackage{flushend}
\usepackage{times}
\usepackage{graphicx}
\usepackage{lipsum}
\usepackage{setspace}
\usepackage{epsfig}
\usepackage{mathptmx}
\usepackage{amsmath}
\usepackage{bm}
\usepackage{amssymb}
\usepackage{tabularx}
\usepackage{mathtools}
\usepackage{color,soul}
\usepackage{paralist}
\usepackage{tikz}
\usepackage{float}
\usepackage{pdfpages}
\usepackage{makecell}
\usepackage{array}
\usepackage{xcolor}
\usepackage{subcaption}
\usepackage{enumerate}
\usepackage{url}
\usepackage{pgfgantt}
\usepackage{amsfonts}
\usepackage{textcomp}
\usepackage{multirow,diagbox}
\usepackage{threeparttable}
\usepackage{booktabs}
\usepackage{bm}

\newcolumntype{L}[1]{>{\raggedright\let\newline\\\arraybackslash\hspace{0pt}}m{#1}}
\newcolumntype{C}[1]{>{\centering\let\newline\\\arraybackslash\hspace{0pt}}m{#1}}
\newcolumntype{R}[1]{>{\raggedleft\let\newline\\\arraybackslash\hspace{0pt}}m{#1}}

\newcommand{\inv}{^{-1}}

\newcommand{\SE}{\mathrm{SE}(3)}

\newcommand{\eqdef}{\vcentcolon=}

\newcommand{\campose}[2]{\prescript{#1}{}{\mathbf{X}}_{#2}}
\newcommand{\objpose}[2]{\prescript{#1}{}{\mathbf{L}}_{#2}}
\newcommand{\worldf}{W}
\newcommand{\cammotion}[3]{\prescript{#1}{#2}{\mathbf{T}}_{#3}}
\newcommand{\objmotion}[3]{\prescript{#1}{#2}{\mathbf{H}}_{#3}}
\newcommand{\objf}{L}
\newcommand{\camf}{X}

\newcommand{\mpoint}[2]{\prescript{#1}{}{\mathbf{m}}_{#2}}

\usepackage{xcolor}  

\def\secref#1{Section~\ref{#1}}
\def\figref#1{Fig.~\ref{#1}}
\def\tabref#1{Table~\ref{#1}}
\def\eqref#1{(\ref{#1})}

\title{\LARGE \bf
The Importance of Coordinate Frames in Dynamic SLAM}

\author{Jesse Morris, Yiduo~Wang and~Viorela~Ila
\thanks{This research is funded with the support of ARIA Research and the Australian Government via the Department of Industry, Science, and Resources CRC-P program (CRCPXI000007).}
\thanks{Jesse Morris, Yiduo Wang and Viorela Ila are with the University of Sydney (USyd), 2006 Sydney, Australia.
{\tt \{jesse.morris,yiduo.wang,viorela.ila\}@sydney.edu.au}}
}

\begin{document}

\maketitle
\thispagestyle{empty}
\pagestyle{empty}

\begin{abstract}
Most Simultaneous localisation and mapping (SLAM) systems have traditionally assumed a static world, which does not align with real-world scenarios. 
To enable robots to safely navigate and plan in dynamic environments, it is essential to employ representations capable of handling moving objects. 
Dynamic SLAM is an emerging field in SLAM research as it improves the overall system accuracy while providing additional estimation of object motions. 
State-of-the-art literature informs two main formulations for Dynamic SLAM, representing dynamic object points in either the world or object coordinate frame. 
While expressing object points in their local reference frame may seem intuitive, it does not necessarily lead to the most accurate and robust solutions. 
This paper conducts and presents a thorough analysis of various Dynamic SLAM formulations, identifying the best approach to address the problem. 
To this end, we introduce a front-end agnostic framework using GTSAM~\cite{gtsam} that can be used to evaluate various Dynamic SLAM formulations.\footnote{Open-source:~\url{https://github.com/ACFR-RPG/dynamic_slam_coordinates}}
\end{abstract}


\section{Introduction}
\label{sec:intro}

Simultaneous localisation and mapping (SLAM) is a problem that has been studied for more than three decades~\cite{Rosen2021annurev}. 
SLAM systems enable robots to create representations of the environment while simultaneously localising themselves within it. 
Many current SLAM solutions operate with the assumption that the environment consists mostly of stationary elements~\cite{Newcombe2011ismar,mur2017orb,Campos2021tro}, 
which may not hold true in real-world situations where dynamic objects are abundant. 

Conventionally, SLAM systems treat sensor data associated with moving objects as outliers
and reject them from the estimation process~\cite{Zhao08icra,Bescos2018ral}, disregarding any useful information pertaining to dynamic objects.
Integrating objects into the SLAM framework has the advantage that the resulting map can directly inform navigation and task planning systems~\cite{Finean2021iros, Hermann2014} of the estimated object motion and scene structure, improving robotic system robustness in complex dynamic environments~\cite{Wang07ijrr, Henein20icra}.
As such, 
an emerging theme in SLAM is to incorporate observations of the dynamic components of the scene and estimate their motions~\cite{Rosen2021annurev} -- in this paper we refer to such a system as Dynamic SLAM.

\begin{figure}[ht]
	\centering
	\includegraphics[width=0.9\columnwidth]{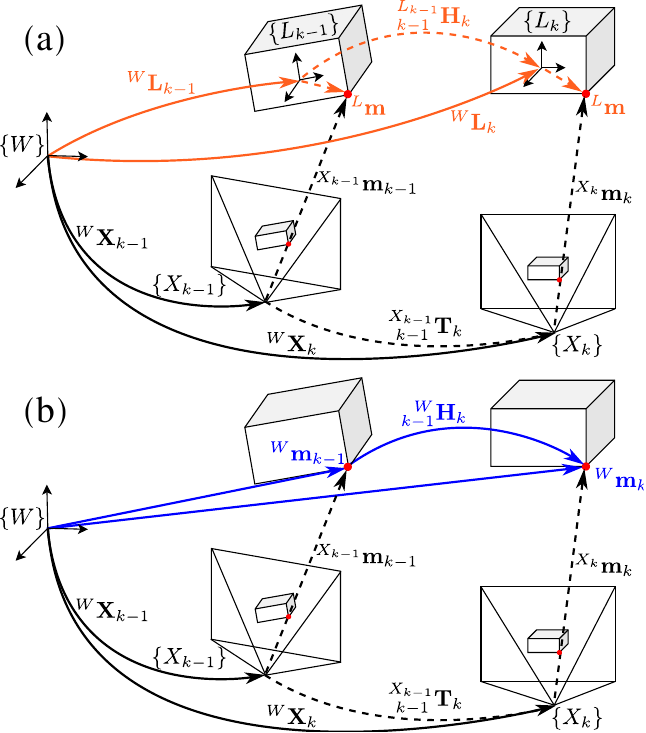}
	\caption{\small{\textbf{Object-centric vs world-centric}. A comparison of two Dynamic SLAM formulations viewing the same scene from time-step $k-1$ to $k$. 
    Three reference frames are included, namely world $\{\worldf\}$, object $\{\objf\}$ and camera $\{\camf\}$. 
    \textbf{(a)} The object-centric formulation expresses dynamic points $\mpoint{\objf}{}$ in the local object frame $\{\objf\}$ defined by an estimate for object pose $\objpose{\worldf}{}$ at each time-step. 
    \textbf{(b)} The world-centric representation describes the rigid body motion $\objmotion{\worldf}{k-1}{k}$ directly with dynamic points $\mpoint{\worldf}{}$.}}
	\label{fig:showcase}
    \vspace{-5mm}
\end{figure}
Recently, multi-object visual odometry techniques~\cite{Judd2020iros, judd2021arxiv} and graph-based optimisation Dynamic SLAM systems~\cite{Bescos2018ral, bescos2021ral, zhang2020vdoslam} have been explored to jointly estimate the robot pose, the static structure and the motion/trajectory of rigid-body objects in the scene based on static and dynamic point observations. 
The literature proposes a variety of ways to formulate this optimisation problem, 
each of which optimises for different sets of variables and objective functions. This defines different underlying graph structures of the optimisation. 
The choice of formulation significantly affects the robustness, accuracy and efficiency of SLAM systems~\cite{Rosen2021annurev}. 
Therefore it is paramount to conduct formal analysis on different formulations to clearly delineate the circumstance that leads to the best performance.

In the context of Dynamic SLAM systems,
the optimisation problem can be formulated by representing variables in different frames of reference. 
A common approach, presented in \figref{fig:showcase} (a), expresses observed dynamic points in the local frame of their corresponding objects, 
which this paper refers to as \textit{object-centric}. 
This method appears intuitive as for rigid bodies, points expressed locally are static with respect to the object frame, allowing each dynamic point to be modelled as a single state variable~\cite{bescos2021ral}. 
As a consequence, the pose of each object per time-step must be a variable in the optimisation problem. 

An alternative approach expresses dynamic points in a known reference frame, 
such as camera~\cite{judd2021arxiv} or map/world frame~\cite{zhang2020vdoslam}.
Our previous work, VDO-SLAM~\cite{Henein20icra, zhang2020vdoslam, zhang20iros}, demonstrates that an $\SE$ motion can be expressed in any reference frame including the world frame. 
With that and by representing dynamic object points in the world frame, \cite{zhang2020vdoslam} produces accurate object motion estimates.
Structurally, this approach results in more state variables as each dynamic point is modelled per time-step, but avoids the need to estimate the pose of each object. 
This paper refers to this formulation as \textit{world-centric}, which is visualised in~\figref{fig:showcase}~(b). 

This paper explores the impact of the formulations resulting from different representations on the underlying optimisation problem, 
so as to understand how to better represent objects in Dynamic SLAM systems. 
To this end, we introduce a graph-based optimisation framework for developing and testing different Dynamic SLAM approaches.
Intrigued by the state-of-the-art literature, we implemented world and object-centric formulations, rigorously analysing the accuracy and robustness of the resulting optimisation problem. 
Based on this analysis, we propose the Dynamic SLAM formulation that most accurately and robustly estimates camera poses and object motions. 

The contributions of this paper are as follows: 
\begin{itemize}
\item introduces a collection of detailed mathematical formulations and graph structures for estimating egomotion and tracking dynamic objects in SLAM problems, 
\item rigorously analyses, evaluates and tests each formulation using real-world datasets 
\item provides a Dynamic SLAM optimisation framework using GTSAM~\cite{gtsam} that implements a variety of formulations as presented in this paper. 
\end{itemize}


\section{Related Work}
\label{sec:related}

Dynamic SLAM is an active area of research in robotics, with several efficient solutions being proposed in recent years~\cite{bescos2021ral,judd2021arxiv,zhang2020vdoslam,Bescos2018ral,hachiuma2019detectfusion,zhang2020flowfusion}. 
Conventional solutions like ORB-SLAM 3~\cite{Campos2021tro} reject dynamic objects as outliers using methods such as RANSAC~\cite{fischler1981cacm}. 
Semantic information from deep learning methods are also used to detect and remove dynamic objects~\cite{Bescos2018ral,hachiuma2019detectfusion,zhang2020flowfusion} to create a global map from which only camera pose and the static structure are estimated. 
These methods can accurately estimate camera pose in dynamic environments; however, relevant information about objects moving in the environment is discarded.

To overcome this problem, recent approaches tightly couple object tracking with SLAM, directly integrating observations of dynamic objects into the SLAM formulation and use joint optimisation methods to provide accurate estimates of the dynamic scene.
These systems rely on separating dynamic points from the static background using either kinematics~\cite{Huang2019iccv, judd2021arxiv} or semantics~\cite{huang2020cvpr, bescos2021ral, zhang2020vdoslam} to model each object individually, 
and optimise the pose or motion of these objects together with the camera/robot locations and the map (e.g. dynamic and static points). 
State-of-the-art literature presents two different solutions to represent the dynamic points, categorised by the reference frames in which these points are expressed. 

The most common and intuitive approach is an object-centric formulation~\cite{bescos2021ral, huang2020cvpr, Ballester2021icra}. The immediate advantage is that each object point can be associated with only one variable in the optimisation problem, reducing the number of variables in the system. 
However, one of the challenges posed by such a formulation is that object poses used as dynamic points' reference frames are not directly observable. 
Among object-centric representations, DynaSLAM II~\cite{bescos2021ral} reports the best egomotion estimation when compared with other Dynamic SLAM approaches. 
Their experimental results present poor object motion estimations and the authors consider their use of sparse features to be the main reason behind such performance~\cite{bescos2021ral}. 

An alternative formulation is to represent dynamic objects and estimate their motions directly in a known reference frame~\cite{Judd18iros, Judd19ral, judd2021arxiv, zhang2020vdoslam}.
In this context, a known frame can either be a camera frame that moves with a sliding window, 
or be a well-defined reference frame, such as the world frame which commonly coincides with the first camera/robot pose. 
MVO~\cite{judd2021arxiv} employs a sliding window to track dynamic objects
and reports accurate camera and object motion estimates. 
Their formulation represents the dynamic points in the camera frame at the start of each sliding window; though it models object motions in the object frame, similar to~\cite{bescos2021ral}. 
MVO uses the object observation at the start of the sliding window as reference. 
Our previous work, VDO-SLAM~\cite{Henein20icra,zhang2020vdoslam, zhang20iros}, proposes a model-free formulation to represent and estimate object motions in any desired reference frame based on the rigid-body assumption. 
VDO-SLAM expresses both dynamic points and object motions in the world frame. 
While this formulation may appear less efficient because it introduces new variables associated with observed dynamic points at each step, our intuition suggests that a world-centric approach could significantly enhance the performance of the nonlinear solver. 

\section{Background}
\label{sec:background}

\subsection{Reference Frames and Notations}
\label{sec:notations}
The particular formulations discussed in this paper are concerned with a robot in motion equipped with an RGB-D camera observing and tracking static and dynamic points in the environment. Robot and camera coordinate frames are assumed to coincide. 
\figref{fig:showcase} presents the basic notations employed by this paper. 
The world frame $\{\worldf\}$ defines the fixed global reference frame. 
Let $\campose{\worldf}{k}, \objpose{\worldf}{k} \in \SE$ be the camera and object poses in $\{\worldf\}$ at time-step $k$, respectively. 
Each $\objpose{W}{k}$ is associated with an object frame $\{\objf_k\}$ and each $\campose{W}{k}$ with a camera frame $\{\camf_k\}$.

Let $\mpoint{}{}^{i} = \left[\tilde{\mathbf{m}}^i, 1\right]^\top$ define the homogeneous coordinates of a 3D point $\tilde{\mathbf{m}}^i\in\mathbb{R}^3$, where $i$ is the unique tracklet index, indicating correspondences between observations. 
A point in the camera frame is denoted as $\mpoint{\camf_k}{k}^{i}$. 
The coordinates of a dynamic point in the world frame $\{\worldf\}$ observed at time $k$ is \mbox{$\mpoint{\worldf}{k}^{i}$}, 
and a static point in the world frame is \mbox{$\mpoint{\worldf}{}^{i} = \campose{\worldf}{k} \: \mpoint{\camf_k}{k}^{i}$}. 
The time-step $k$ is omitted when the variable is time-independent, i.e. static, within the represented reference frame. 
A point in object frame $\{\objf_k^j\}$ is $\mpoint{\objf^j}{}^{i}$ where $j$ is a unique object identifier. 
The same point can be expressed in $\{\worldf\}$ as $\mpoint{\worldf}{k}^{i} = \objpose{\worldf}{k}^j \: \mpoint{\objf^j}{}^{i}$, where the $j$ becomes implicit.

\figref{fig:showcase} further highlights how this notation extends to homogeneous transformations. 
$\cammotion{\camf_{k-1}}{k-1}{k} \in \SE$ describes the relative camera transformation from time-step $k-1$ to $k$, expressed in the camera frame $\{\camf_{k-1}\}$, 
and $\objmotion{\objf_{k-1}}{k-1}{k}^{j} \in \SE$ describes the motion for object $j$ in the object frame $\{\objf_{k-1}^j\}$:
\begin{align}
\label{eq:motion_cam_in_L}
    \cammotion{\camf_{k-1}}{k-1}{k} &= \campose{\worldf}{k-1}^{-1}\: \campose{\worldf}{k} \\
\label{eq:motion_object_in_L}
    \objmotion{\objf_{k-1}}{k-1}{k}^{j} &= \objpose{\worldf}{k-1}^{j\ -1}\: \objpose{\worldf}{k}^{j}\text{,}
\end{align}
defining the kinematic models for camera and object. 

\subsection{Pose Transformation and Frame Change}
\label{sec:frame_change}
Our previous work~\cite{zhang2020vdoslam} demonstrates that, 
for a rigid-body object $j$ with motion $\objmotion{\objf_{k-1}}{k-1}{k}^j$, 
there exists a single $\SE$ transformation from time-step $k-1$ to $k$ for all points on this object in the world frame $\{\worldf\}$:
\begin{equation}
\begin{aligned}
	\label{eq:motion_point_rigid}
	\mpoint{\worldf}{k}^{i} =& \objpose{\worldf}{k-1}^j \:
	\objmotion{\objf_{k-1}}{k-1}{k}^j \: \objpose{\worldf}{k-1}^{j\ -1} \: \mpoint{\worldf}{k-1}^{i} \\
	\mpoint{\worldf}{k}^{i} =& 
	\objmotion{\worldf}{k-1}{k}^j \: \mpoint{\worldf}{k-1}^{i}\text{,}
\end{aligned}
\end{equation}
where $\objmotion{\worldf}{k-1}{k}^j$ describes the motion of a point on a rigid-body. 
\begin{equation}
\begin{aligned}
	\label{eq:motion_frame_transform}
	\objmotion{\worldf}{k-1}{k}^j \eqdef& 
	\objpose{\worldf}{k-1}^j \: \objmotion{\objf_{k-1}}{k-1}{k}^j \: \objpose{\worldf}{k-1}^{j\ -1} \in \SE\\
\end{aligned}
\end{equation}
Equation~\eqref{eq:motion_frame_transform} represents a \emph{frame change of a pose transformation}~\cite{Chirikjian17idetc}, relating $\objmotion{\objf_{k-1}}{k-1}{k}^j$, the motion in the object (or body) frame, to that in a world (inertial) reference frame $\objmotion{\worldf}{k-1}{k}^j$.
Using a world reference frame allows the object motion to be described in a model-free manner, eliminating the need to consider the object pose in the formulation.
Based on~\eqref{eq:motion_object_in_L} and~\eqref{eq:motion_point_rigid}, 
the kinematic model that describes the object motion in the world frame $\{\worldf\}$ is as follows:
\begin{equation}
\label{eq:motion_kinematic_constraint}
    \objmotion{W}{k-1}{k}^{j} =  \objpose{W}{k}^{j} \: \objpose{W}{k-1}^{j\ -1} \in \SE\text{.}
\end{equation}

\section{Formulations}
\label{sec:methods}
This section introduces several formulations to define variables and model relations (factors) between those variables in a factor-graph-based Dynamic SLAM estimation framework similar to state-of-the-art approaches~\cite{bescos2021ral, judd2021arxiv, zhang2020vdoslam}. 
We categorise these formulation as either world-centric (\secref{sec:world_centric}) or object-centric (\secref{sec:object_centric}). 

\subsection{SLAM front-end}
\label{sec:front-end}
This paper focuses on the factor-graph-based Dynamic SLAM optimisation (e.g. back-end or local batch) and the proposed framework is intended to be front-end-agnostic. 
The discussion on a complete Dynamic SLAM pipeline or its front-end is not within the scope of this paper. 
The interface between the front-end and the back-end is streamlined so that the front-end can be easily replaced. 

The front-end is expected to provide frame-to-frame tracking for all (static and dynamic) 3D points $\mpoint{\camf_k}{}^{i}$ and to be able to associate/cluster dynamic points by the corresponding objects. 
Furthermore, it can also provide initial estimates for camera poses $\campose{W}{k}$ and object motion $\objmotion{W}{k}{k-1}^{j}$ used to track the static and dynamic points.

\subsection{World-centric Formulation}
\label{sec:world_centric}
\begin{figure}[t]
	\centering
	\includegraphics[trim={0cm 0cm 0.8cm 0cm},clip,width=0.89\columnwidth]{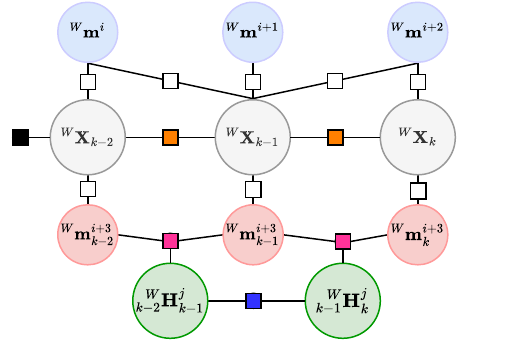}
	\caption{\small{\textbf{World-centric formulation factor graph}. 
    Example factor graph including three static points $\mpoint{\worldf}{}^{i:i+2}$ and one dynamic point $\mpoint{\worldf}{}^{i+3}$ on object $j$ seen at three consecutive time-steps $k-2:k$. 
    Point measurement factors are shown as white squares, odometry factors as orange squares and world-centric motion factors as magenta squares. 
    The motion smoothing factor is shown in blue and the prior factor is in black.}}
    \label{fig:world_centric_graph}
    \vspace{-6mm}
\end{figure}

The world-centric formulation jointly estimates for camera pose, object motion, static and dynamic points, all expressed in the world frame $\{\worldf\}$~\cite{zhang2020vdoslam}. 
A conceptually similar variant would represent both in the first camera of a sliding window optimisation problem.
\figref{fig:world_centric_graph} shows a simple example of the corresponding factor graph. 

Given an observation of a 3D point $\mpoint{\camf_k}{k}^{i}$, the \emph{point measurement factor} models the camera pose $\campose{\worldf}{k}$ with a map point $\mpoint{\worldf}{k}^{i}$ and is given by:
\begin{equation}
\label{eq:world_point3d_factor}
    r(\campose{\worldf}{k}, \mpoint{\worldf}{}^{i}) = \mpoint{\camf_k}{k}^{i} - \campose{\worldf}{k}^{\inv} \: \mpoint{\worldf}{}^{i}\text{,}
\end{equation} 
where $\campose{\worldf}{k}$ and $\mpoint{\worldf}{}^{i}$ are vertices in the factor graph and require initialisation.
The initial value for $\campose{\worldf}{k}$ is provided by the front-end, 
and $\mpoint{\worldf}{}^{i}$ is initialised as $\mpoint{\worldf}{}^{i} = \campose{\worldf}{k} \: \mpoint{\camf_k}{k}^{i}$. 
As shown in~\figref{fig:world_centric_graph}, the same factor is used to refine dynamic points $\mpoint{\worldf}{k}^{i}$ as well.

The \emph{camera odometry factor} between consecutive camera poses in the graph is formulated as:
\begin{equation}
\label{eq:world_cam_odom_factor}
    r(\campose{\worldf}{k-1},\campose{\worldf}{k}) = \left[\log\left(\campose{\worldf}{k}^{\inv} \: \campose{\worldf}{k-1} \: \cammotion{\camf_{k-1}}{k-1}{k}\right)\right]^{\vee}\text{,}
\end{equation}
where the relative pose change $\cammotion{\camf_{k-1}}{k-1}{k}$ is given by the front-end.
The operation $\left[\log\left(\cdot\right)\right]^{\vee}$ maps an $\SE$ transformation to an $\mathbb{R}^6$ vector as per the notations of Chirikjian~\cite{Chirikjian1994}. 

Based on~\eqref{eq:motion_point_rigid}, 
the motion of a point on a rigid body is described by a ternary motion factor, relating a pair of tracked points with their motion:
\begin{equation}
\label{eq:world_landmark_motion_tenary_factor}
    r(\mpoint{\worldf}{k}^{i}, \mpoint{\worldf}{k-1}^{i}, \objmotion{\worldf}{k-1}{k}^{j}) = \mpoint{\worldf}{k}^{i} - \objmotion{\worldf}{k-1}{k}^{j} \: \mpoint{\worldf}{k-1}^{i}\text{.}
\end{equation}
In~\eqref{eq:world_landmark_motion_tenary_factor}, the points from tracklet $i$ are on the $j$-th object and observed at time-step $k-1$ and $k$, forming the \emph{world-centric motion factors} in~\figref{fig:world_centric_graph}. 

Finally, 
the \emph{smoothing factor} is introduced between consecutive object motions:
\begin{equation}
\label{eq:world_motion_smoothing_factor}
    r(\objmotion{\worldf}{k-2}{k-1}^{j},  \objmotion{\worldf}{k-1}{k}^{j}) =  \left[\log\left(\objmotion{\worldf}{k-2}{k-1}^{j\ -1} \: \objmotion{\worldf}{k-1}{k}^j\right)\right]^{\vee}\text{.}
\end{equation}
This factor helps prevent abrupt, drastic and unrealistic changes in object motions between consecutive frames. 

\subsection{Object-centric Formulation}
\label{sec:object_centric}
\begin{figure}[t]
	\centering
	\includegraphics[trim={0cm 0cm 0cm 0cm},clip,width=0.8\columnwidth]{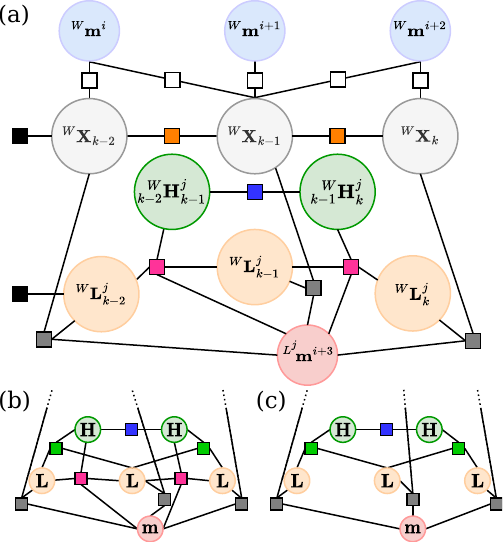}
	\caption{\small{\textbf{Object-centric formulation factor graph}. 
    Example factor graphs showing three static points $\mpoint{\worldf}{}^{i:i+2}$ and a tracked point $\mpoint{\objf^j}{}^{i+3}$ on a dynamic object $\objpose{}{}^j$, seen at consecutive time-steps $k-2:k$. 
    Dynamic point measurement factors are shown as gray squares, object-centric motion factors are magenta squares and green squares represent object kinematic factors. The motion smoothing factor is blue and priors factors are black. (b) and (c) show variations of the graph in (a).}}
    \label{fig:object_centric_graph}
    \vspace{-6mm}
\end{figure}

The object-centric approach estimates camera pose, static points, object motion and pose in $\{\worldf\}$, and object points in $\{\objf\}$.
The corresponding factor graph is visualised in~\figref{fig:object_centric_graph} which shows different object-centric variations used for experiments.
Variation (a) shows the basic object-centric structure, 
and (b) modifies the graph structure to include the \emph{object kinematic factor} (\secref{sec:object_kinematic_factor}), while (c) retains this factor and removes the \emph{object-centric motion factor}. 
To ensure a fair comparison with the world-centric formulation, we retain common factors where possible, i.e. point measurement factors for static points and odometry factors, 
as indicated by identical connections between~\figref{fig:world_centric_graph} and~\figref{fig:object_centric_graph}. 
These variations are introduced so that we can explore the effect that different object-centric factors have on the underlying optimisation structure, behaviour and performance. 

Equation~\eqref{eq:world_point3d_factor} is extended to express dynamic points in object frame, additionally constraining the object pose:
\begin{equation}
\label{eq:object_point3d_factor}
    r(\campose{\worldf}{k}, \objpose{\worldf}{k}^{j}, \mpoint{\objf}{}^{i}) = \mpoint{\camf_k}{k}^{i} - \campose{\worldf}{k}^{\inv} \: \objpose{\worldf}{k}^{j} \: \mpoint{\objf}{}^{i}\text{.}
\end{equation}
Following the rigid body assumption we consider points in object frame $\mpoint{\objf}{}$ to be time independent -- they are static relative to the object frame $\{\objf\}$.

At each time-step, the translation component of the object pose $\objpose{\worldf}{k}^{j}$ is initialised using the centroid of the tracked object points and the rotation component is initialised with identity matrix~\cite{bescos2021ral}. 
This initial object pose is used to initialise each new dynamic point first seen at that time step: $\mpoint{\objf^j}{}^i = \objpose{\worldf}{k}^{j\ -1} \: \campose{\worldf}{k} \: \mpoint{\camf_k}{k}^i$. 

The \emph{object-centric motion factor} now connects points $\mpoint{\objf^j}{}$, consecutive object poses $\objpose{\worldf}{}^j$, and object motion $\objmotion{\worldf}{}{}^j$:
\begin{equation}
\begin{aligned}
    \label{eq:object_landmark_motion_factor}
   &r(\objpose{\worldf}{k}^j, \objpose{\worldf}{k-1}^j, \objmotion{\worldf}{k-1}{k}^j, \mpoint{\objf^j}{}^i) \\
   =&\objpose{\worldf}{k}^j \: \mpoint{\objf^j}{}^i - \objmotion{\worldf}{k-1}{k}^j \: \objpose{\worldf}{k-1}^j \: \mpoint{\objf^j}{}^i \\ 
   =&\left(\objpose{\worldf}{k}^j - \objmotion{\worldf}{k-1}{k}^j \: \objpose{\worldf}{k-1}^j\right)\mpoint{\objf^j}{}^i\text{.}
\end{aligned}
\end{equation}
As this residual is the only factor containing $\objmotion{}{}{}$, it should encode the kinematic model that uses object pose to define the object motion and the motion of a point on rigid body as expressed in~\eqref{eq:motion_kinematic_constraint} and~\eqref{eq:motion_point_rigid}, respectively.
However, this factor does not actually reflect the kinematic model established in~\eqref{eq:motion_kinematic_constraint} as \mbox{$\objpose{\worldf}{k}^j - \objmotion{\worldf}{k-1}{k}^j \: \objpose{\worldf}{k-1}^j \notin \SE$}.

\subsection{Object Kinematic Factor}
\label{sec:object_kinematic_factor}
We therefore propose adding an additional factor to directly model the kinematic relationship between consecutive object poses:
\begin{equation}
\label{eq:object_kinematic_factor}
   r(\objpose{\worldf}{k}^j, \objpose{\worldf}{k-1}^j, \objmotion{\worldf}{k-1}{k}^j) = \left[\log\left(\objpose{\worldf}{k}^{j\ -1} \:  \objmotion{\worldf}{k-1}{k}^j \: \objpose{\worldf}{k-1}^j\right)\right]^{\vee}
\end{equation}
We refer to it as the \textit{object kinematic factor} and is shown as green squares in~\figref{fig:object_centric_graph} (b) and (c). It explicitly describes the change in object pose $\objpose{\worldf}{}$ between time-step $k-1$ and $k$ with an object motion $\objmotion{\worldf}{k-1}{k}$ derived from~\eqref{eq:motion_kinematic_constraint}. 

\begin{table*}[ht]
\footnotesize
\centering
\setlength{\tabcolsep}{12pt}
\caption{\small{Camera pose estimation (mean RPE, where $\mathbf{M} = \campose{\worldf}{k-1}^{-1}\campose{\worldf}{k}$) on KITTI sequences~\cite{Geiger12cvpr} comparing DynaSLAM II~\cite{bescos2021ral} with object and world-centric formulations. The number of variables in the factor graph (\# var) for object-centric and world-centric formulations are additionally included, as well as the time to solve the full Dynamic SLAM system, which is averaged over $10$ runs.}}
\label{tab:kitti_cam}
\begin{tabular}{c|cc|cc|cc|cc}
\toprule
 & 
 \multicolumn{2}{c|}{DynaSLAM II} & 
 \multicolumn{2}{c|}{\makebox[0pt]{object-centric}} & 
 \multicolumn{2}{c|}{\makebox[0pt]{object-centric with OKF}} & 
 \multicolumn{2}{c}{\makebox[0pt]{world-centric}} \\
\multirow{1}{*}{\makebox[0pt]{Seq}} & $E_r$(\si{\degree}) & $E_t$(m) & $E_r$(\si{\degree}) &$E_t$(m) & $E_r$(\si{\degree}) &$E_t$(m) & $E_r$(\si{\degree}) &$E_t$(m) \\
\midrule
\midrule
\multirow{1}{*}{\makebox[0pt]{00}} & $0.06$ & $\mathbf{0.04}$ & $0.06$ & $0.05$  & $0.09$ & $0.05$ & $\mathbf{0.05}$ & $\mathbf{0.04}$ \\
\multirow{1}{*}{\makebox[0pt]{01}} & $\mathbf{0.04}$ & 0.05 & 0.05 & 0.04  & 0.07 & 0.04 & $\mathbf{0.04}$ & $\mathbf{0.03}$ \\
\multirow{1}{*}{\makebox[0pt]{02}} & $\mathbf{0.02}$ & 0.04 & $\mathbf{0.02}$ & 0.03 & 0.06 & 0.03 & $\mathbf{0.02}$ & $\mathbf{0.03}$ \\
\multirow{1}{*}{\makebox[0pt]{03}} & $0.04$ & $\mathbf{0.06}$ & $0.05$ & $0.07$ & $0.09$ & $0.07$ & $\mathbf{0.03}$ & $\mathbf{0.06}$ \\
\multirow{1}{*}{\makebox[0pt]{04}} & $0.06$ & $0.07$ & $\mathbf{0.04}$ & $0.07$ & $0.06$ & $0.06$  & $\mathbf{0.04}$ & $\mathbf{0.06}$ \\
\multirow{1}{*}{\makebox[0pt]{05}} & $0.03$ & $\mathbf{0.06}$ & $0.03$ & $\mathbf{0.06}$ & $0.07$ & $\mathbf{0.06}$ & $\mathbf{0.02}$ & $\mathbf{0.06}$ \\
\multirow{1}{*}{\makebox[0pt]{06}} & $\mathbf{0.04}$ & $0.02$ & $0.06$ & $0.02$ & $0.07$ & $0.02$ & $0.05$ & $\mathbf{0.01}$ \\
\multirow{1}{*}{\makebox[0pt]{18}} & $\mathbf{0.02}$ & $0.05$ & $0.03$ & $\mathbf{0.04}$ & $0.1$ & $\mathbf{0.04}$ & $\mathbf{0.02}$ & $\mathbf{0.04}$ \\ 
\multirow{1}{*}{\makebox[0pt]{20}} & $0.04$ & $0.07$ & $\mathbf{0.03}$ & $\mathbf{0.05}$ & $0.1$ & $\mathbf{0.05}$ & $\mathbf{0.03}$ & $\mathbf{0.05}$ \\
\bottomrule
\end{tabular}
\begin{tabular}{cc|cc}
\toprule
\multicolumn{2}{c|}{\makebox[0pt]{object-centric}}  & 
\multicolumn{2}{c}{\makebox[0pt]{world-centric}} \\
\multirow{1}{*}{\makebox[10pt]{\# var}} & \multirow{1}{*}{\makebox[10pt]{time(\si{\second})}} & \multirow{1}{*}{\makebox[10pt]{\# var}} & \multirow{1}{*}{\makebox[10pt]{time(\si{\second})}} \\   
\midrule
\midrule
\makebox[0pt]{26335} & \makebox[0pt]{$365.3$} & \makebox[0pt]{153155} & \makebox[0pt]{$\mathbf{72.0}$} \\
\makebox[0pt]{51426} & \makebox[0pt]{$38.4$} & \makebox[0pt]{117923} &  \makebox[0pt]{$\mathbf{34.2}$} \\
\makebox[0pt]{19034} & \makebox[0pt]{$102.0$} & \makebox[0pt]{38450} & \makebox[0pt]{$\mathbf{22.4}$} \\
\makebox[0pt]{20410} & \makebox[0pt]{$215.4$} & \makebox[0pt]{92264} & \makebox[0pt]{$\mathbf{24.8}$} \\
\makebox[0pt]{36486} & \makebox[0pt]{$39.8$} & \makebox[0pt]{80352} & \makebox[0pt]{$\mathbf{22.3}$} \\
\makebox[0pt]{31369} & \makebox[0pt]{$63.9$} & \makebox[0pt]{99990} & \makebox[0pt]{$\mathbf{41.8}$} \\
\makebox[0pt]{21353} & \makebox[0pt]{$225.7$} & \makebox[0pt]{91187} & \makebox[0pt]{$\mathbf{73.9}$} \\
\makebox[0pt]{53680} & \makebox[0pt]{$\mathbf{269.2}$} & \makebox[0pt]{340844} & \makebox[0pt]{$381.5$} \\
\makebox[0pt]{129316} & \makebox[0pt]{$\mathbf{422.6}$} & \makebox[0pt]{711804} & \makebox[0pt]{$656.9$} \\
\bottomrule
\end{tabular}
\vspace{-5mm}
\end{table*}

\section{Experiments}
\label{sec:expe}
The formulations presented in~\secref{sec:methods} are implemented and optimised using GTSAM~\cite{gtsam}. 
The multi-motion visual odometry component of our previous work~\cite{zhang2020vdoslam} is used as the front-end to provide  
frame-to-frame tracking and initial estimations for points, camera poses and object motions.

For each experiment, the front-end output is saved as a graph file~\cite{Kummerle11icra} to provide identical data association, measurements and initial estimates, eliminating variation and randomness to ensure consistent input for each test. 
The same input measurements and initial estimates are used to construct each system, depending on the desired graph structure and reference frame.

The KITTI Tracking Dataset~\cite{Geiger13ijrr} is used to assess the performance of each formulation. 
Sequences with a sufficient variety of dynamic objects are selected,
as some sequences contain no moving objects, 
or objects with very short trajectories. 
For each selected sequence, 
we evaluate the solution accuracy and analyse the behaviour of the optimisation for all world and object-centric formulations. 
The effect of different object-centric factors is investigated by including the object-centric variations (\figref{fig:object_centric_graph}) in our experiments.

For comparison, we further include the camera pose errors of DynaSLAM II~\cite{bescos2021ral}, the state-of-the-art in egomotion estimation, as reported in their paper since DynaSLAM II is not open-source.
However, we omit their object motion error in our comparison because their object motion estimation~\cite{bescos2021ral} performs comparably to our object-centric formulation, and their error metrics are not specified. 
MVO~\cite{judd2021arxiv}, another closed-source system, is formulated similarly to the world-centric approach presented in this paper. 
However, their system uses a sliding window optimisation; therefore, their results are not comparable.

\subsection{Error Metrics}
\label{sec:error-metrics}
The paper reports Relative Pose Error (RPE)~\cite{sturm2012iros} for both camera and objects computed as follows.
Given a ground truth transformation $\mathbf{M}_\text{gt}\in \SE$ and a corresponding estimate $\mathbf{M}\in \SE$, we compute the error as $\mathbf{E} = \mathbf{M}^{-1} \:  \mathbf{M}_\text{gt}$ for all $\SE$ estimates.
The translational error ${E}_{t}$ is the $L_2$ norm of the translational component of $\mathbf{E}$, 
and the rotational error ${E}_{r}$ is the angle of its rotational component. Each table will indicate what transformation $\mathbf{M}$ represents.

\subsection{Camera Pose Error \& Factor Graph}

\tabref{tab:kitti_cam} shows the evaluation results of estimated camera poses from the different formulations.
The world-centric formulation provides the best results among all methods in the most sequences, 
consistently performing better than, or at least on a par with, the state-of-the-art benchmark. 
However, all formulations present similar accuracy for camera pose estimations with minor differences. 
We believe that it is because there are many static background features throughout KITTI sequences to enable accurate camera tracking.

The number of variables in each formulation and associated optimisation time are presented in~\tabref{tab:kitti_cam}. 
Despite a greater number of variables, the world-centric approach on average takes substantially less time to provide a solution while producing more accurate object motion and pose estimates, as shown in~\tabref{tab:kitti_obj_motion} and~\ref{tab:kitti_obj_pose}. 
This highlights how the graph structures resulting from different representation choices have a clear impact on the optimisation efficiency. 

\begin{table}[t]
\footnotesize
\centering
\setlength{\tabcolsep}{4.8pt}
\caption{\small{Errors of object motion, $\objmotion{}{}{}$, on sequences with well-tracked dynamic objects, with $\mathbf{M} = \objmotion{\worldf}{k-1}{k}$. The average error for dynamic objects tracked for the most frames are included, as is the mean error across all objects in the sequence. The object-centric variations correspond to the factor graphs visualised in~\figref{fig:object_centric_graph} (a), (b) and (c) respectively. Blue entries represent the best object-centric estimations, and the bold ones are the best results overall.}}
\label{tab:kitti_obj_motion}
\begin{tabular}{c|cc|cc|cc|cc}
\toprule
 & 
 \multicolumn{2}{c|}{\makebox[0pt]{object-centric}} & 
 \multicolumn{2}{c|}{\makebox[0pt]{object-centric}} & 
 \multicolumn{2}{c|}{\makebox[0pt]{object-centric}} & 
 \multicolumn{2}{c}{\makebox[0pt]{world-centric}} \\
 & & & 
 \multicolumn{2}{c|}{\makebox[0pt]{with OKF}} & 
 \multicolumn{2}{c|}{\makebox[0pt]{only OKF}} & \\
\multirow{1}{*}{\makebox[0pt]{Seq\textbar obj}} & \makebox[10pt]{$E_r$(\si{\degree})} & \makebox[10pt]{$E_t$(m)} & \makebox[10pt]{$E_r$(\si{\degree})} & \makebox[10pt]{$E_t$(m)} & \makebox[10pt]{$E_r$(\si{\degree})} & \makebox[10pt]{$E_t$(m)} & \makebox[10pt]{$E_r$(\si{\degree})} & \makebox[10pt]{$E_t$(m)} \\
\midrule
\midrule
00\textbar 01  & $3.54$ & $1.7$ & $1.27$ & $0.49$ & \textcolor{blue}{$1.01$} & $\mathbf{0.4}$ & $\mathbf{0.9}$ & $0.51$ \\
mean & $4.73$ & $3.85$ & $1.67$ & $1.35$ & \textcolor{blue}{$1.05$} & \textcolor{blue}{$1.11$} & $\mathbf{0.78}$ & $\mathbf{0.52}$ \\
\midrule
03\textbar 01 & $2.14$ & $1.07$ & $3.87$ & $1.26$ & \textcolor{blue}{$0.68$} & \textcolor{blue}{$0.4$} & $\mathbf{0.22}$ & $\mathbf{0.17}$ \\
mean & $0.93$ & $5.39$ & $2.38$ & $1.12$ & \textcolor{blue}{$0.46$} & \textcolor{blue}{$0.34$} & $\mathbf{0.24}$ & $\mathbf{0.23}$  \\
\midrule
04\textbar 03 & $5.22$ & $2.47$ & $2.7$ & $1.08$ &  \textcolor{blue}{$1.82$} &  \textcolor{blue}{$0.73$} & $\mathbf{0.64}$ & $\mathbf{0.37}$ \\
04\textbar 04 &  \textcolor{blue}{$1.13$} & $4.63$ & $1.9$ & $0.85$ & $1.19$ &  \textcolor{blue}{$0.59$} & $\mathbf{0.38}$ & $\mathbf{0.23}$ \\
04\textbar 05 & $3.57$ & $6.55$ & $2.56$ & $1.21$ &  \textcolor{blue}{$1.13$} &  \textcolor{blue}{$0.62$} & $\mathbf{0.5}$ & $\mathbf{0.33}$ \\
mean & $5.48$ & $5.08$ & $2.13$ & $2.11$ &  \textcolor{blue}{$1.81$} &  \textcolor{blue}{$1.35$} & $\mathbf{0.77}$ & $\mathbf{0.51}$ \\
\midrule
05\textbar 20 &  \textcolor{blue}{$1.91$} & $6.54$ & $4.79$ & $30.77$ & $1.94$ &  \textcolor{blue}{$5.38$} & $\mathbf{0.53}$ & $\mathbf{1.37}$ \\
05\textbar 24 & $4.74$ & $18.28$ &  \textcolor{blue}{$1.19$} &  \textcolor{blue}{$5.48$} & $3.45$ & $15.04$ & $\mathbf{0.51}$ & $\mathbf{2.28}$ \\
mean & $1.66$ & $13.68$ & $\mathbf{0.18}$ & $12.43$ & $2.01$ &  \textcolor{blue}{$7.73$} & $0.7$ & $\mathbf{5.19}$\\
\midrule
18\textbar 04 & $6.1$ & $14.50$ & $1.83$ & $4.68$ &  \textcolor{blue}{$0.89$} &  \textcolor{blue}{$4.07$} & $\mathbf{0.25}$ & $\mathbf{1.82}$ \\
mean & $\mathbf{0.25}$ & $11.26$ & $0.97$ & $25.67$ & $1.17$ &  \textcolor{blue}{$4.67$} & $0.52$ & $\mathbf{1.95}$\\
\midrule
20\textbar 32 & $3.60$ & $2.17$ & $0.42$ & $0.23$ &  \textcolor{blue}{$0.32$} &  \textcolor{blue}{$0.17$} & $\mathbf{0.08}$ & $\mathbf{0.15}$ \\
mean & $4.90$ & $14.70$ &  \textcolor{blue}{$1.15$} &  \textcolor{blue}{$5.96$} & $1.21$ & $7.17$ & $\mathbf{0.69}$ & $\mathbf{5.46}$\\
\bottomrule
\end{tabular}
\vspace{-6mm}
\end{table}

\subsection{Object Motion Error}
\label{sec:object_motion_error}
Object motion errors are shown in \tabref{tab:kitti_obj_motion}.
The world-centric formulation is the most accurate overall, outperforming the state-of-the-art, i.e. the object-centric formulation, for \SI{\sim 95}{\percent} of dynamic objects. 
Our proposed object-centric variations improve the accuracy of the state-of-the-art, but the world-centric method still produces the best motion estimations in \SI{\sim 80}{\percent} of all objects. 
The poor results of the base object-centric approach suggests that the object-centric motion factor is unable to effectively contribute to the optimisation as the kinematic model is not correctly encoded.

To further probe this observation, the object kinematic factor (OKF) described in~\eqref{eq:object_kinematic_factor} is subsequently included in the optimisation. 
These results are denoted in all tables as `object-centric with OKF'. 
This factor explicitly enforces the kinematic model in~\eqref{eq:motion_kinematic_constraint}, and improves the motion estimate substantially, particularly in translation.

\begin{figure}[t]
\centering
\begin{subfigure}{0.489\textwidth}
    \centering
   \includegraphics[trim={3cm 0.5cm 3cm 0.5cm},clip,width=1.0\columnwidth]{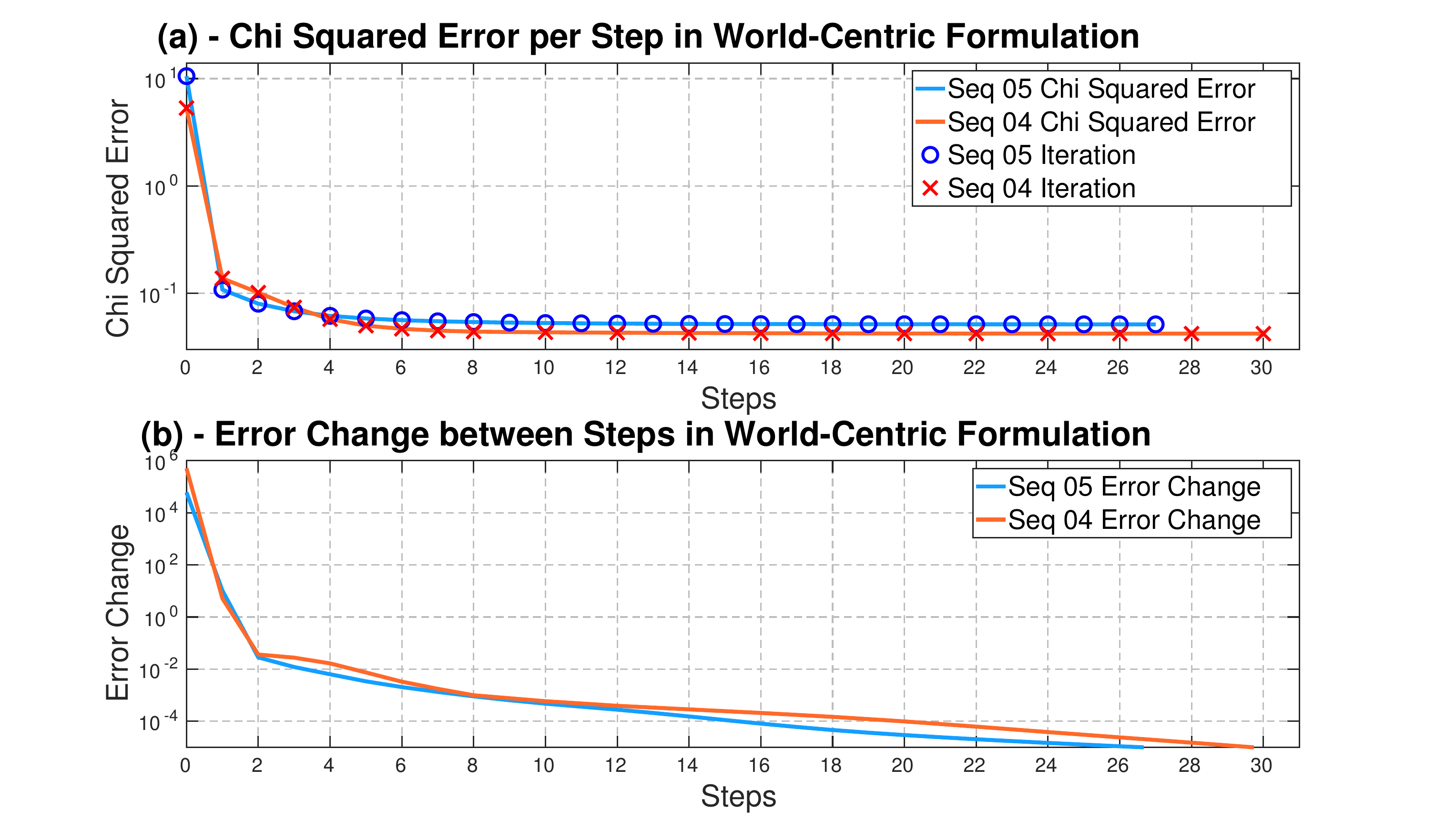}
\end{subfigure}
\begin{subfigure}{0.489\textwidth}
    \centering
   \includegraphics[trim={3cm 1.0cm 3cm 0.5cm},clip,width=1.0\columnwidth]{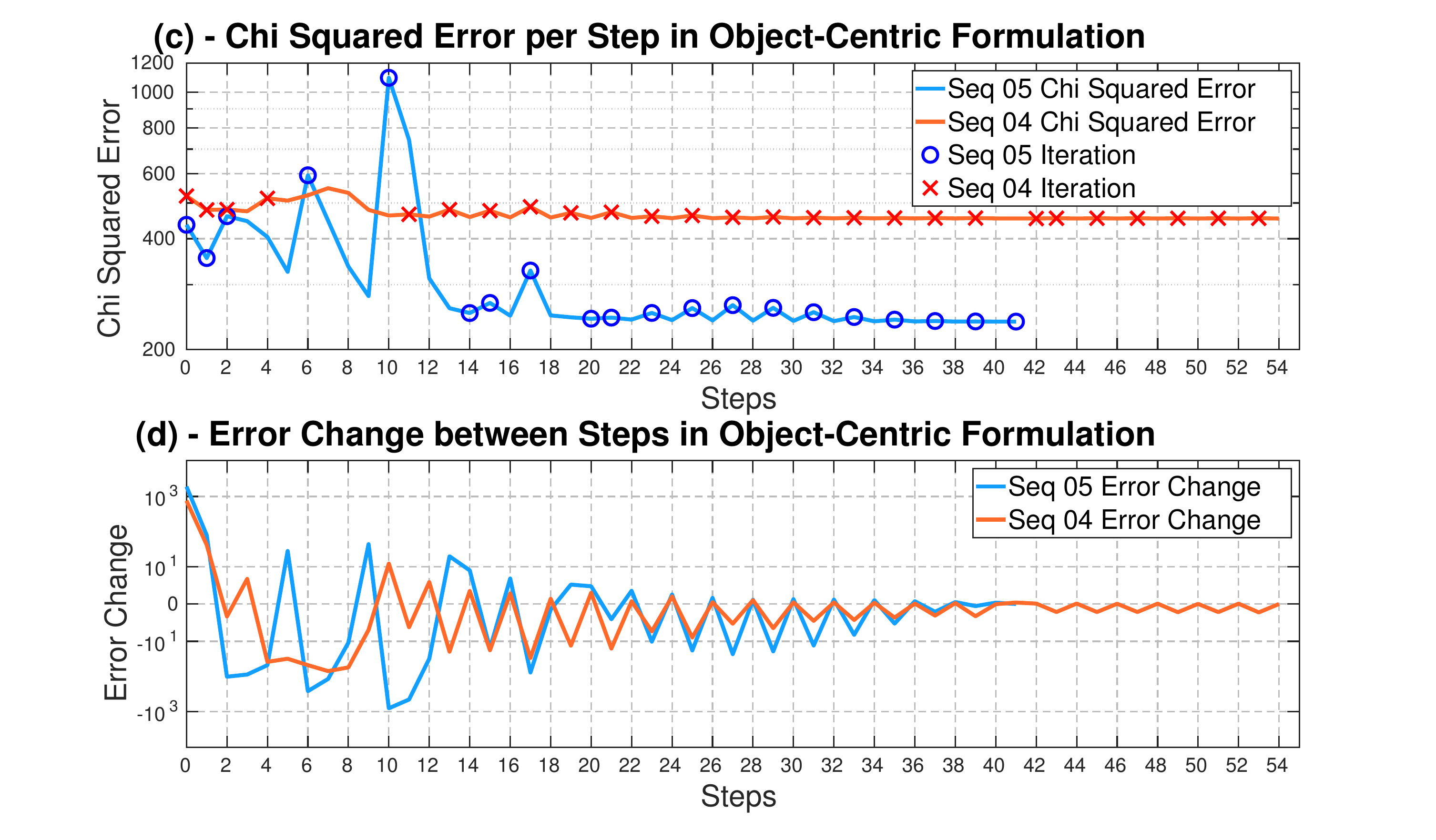}
\end{subfigure}
\vspace{-5mm}
\caption{\small{\textbf{Chi-squared errors $\chi_{n}$ for world and object-centric formulations per-step during optimisation for sequences 04 and 05}. The Error Change is computed as $\chi_{n-1}^2 - \chi_{n}^2$ where $n$ represents the step. The linear system is solved at every step but only re-linearised every iteration, which is marked in plots (a) and (c).}}
\label{fig:optimisation_errors}
\vspace{-6mm}
\end{figure}

Our results indicate that object-centric motion factor alone is detrimental to the optimisation problem as it `pulls' the optimisation in several directions, resulting in sub-optimal solutions. 
We further investigated and analysed the evolution of the chi-squared errors of the world and object-centric formulations during nonlinear least squares optimisation using Levenberg-Marquardt (LM) solver, as per~\figref{fig:optimisation_errors}. 
The x-axis denotes the number of steps that the LM solver requires for convergence and indicates the amount of times a linear system is solved. 
The world-centric system in~\figref{fig:optimisation_errors} (a) exhibits a consistent downward trend in the chi-squared error, 
while the per-step error change of the object-centric formulation, shown in~\figref{fig:optimisation_errors} (c), oscillates between positive and negative, requiring more steps than the former. 
We believe this is the reason behind the poor efficiency reported in~\tabref{tab:kitti_cam}. 
While not presented due to limited space, the convergence trends displayed in~\figref{fig:optimisation_errors} are common to all sequences and will be included in the supplementary material. 

Removing the object-centric motion factor improves the overall estimation accuracy, as shown in the `object-centric only OKF' columns of~\tabref{tab:kitti_obj_motion} and~\ref{tab:kitti_obj_pose}, further emphasising the detrimental effect of this factor.
While this formulation exhibits better performance than the other object-centric variations, only retaining the object kinematic factor limits the effectiveness of the SLAM problem as the point measurements are no longer used to model the rigid-body point motion expressed in~\eqref{eq:motion_point_rigid}.
We additionally noted that the object centric formulations require an extra prior on the first pose of each object trajectory to avoid an indeterminate linear system  --- this was also observed in~\cite{bescos2021ral}.

In contrast, the world-centric formulation does not require an object prior, and explicitly models the rigid-body motion only using variables in a known reference frame $\{\worldf\}$. The optimisation problem arising from this formulation results in a more stable optimisation process, as shown in~\figref{fig:optimisation_errors} (a) and (b), and the most accurate estimation in our experiments. 

\subsection{Object Pose Error}

\begin{table}
\footnotesize
\centering
\setlength{\tabcolsep}{4.8pt}
\caption{\small{Relative errors for object pose, $\objpose{}{}$, with $\mathbf{M} = \objpose{\worldf}{k-1}^{-1}\: \objpose{\worldf}{k}$. Blue entries are the best object-centric estimations, and the bold ones are the best results overall.}}
\label{tab:kitti_obj_pose}
\begin{tabular}{c|cc|cc|cc|cc}
\toprule
 & 
 \multicolumn{2}{c|}{\makebox[0pt]{object-centric}} & 
 \multicolumn{2}{c|}{\makebox[0pt]{object-centric}} & 
 \multicolumn{2}{c|}{\makebox[0pt]{object-centric}} & 
 \multicolumn{2}{c}{\makebox[0pt]{world-centric}} \\
 & & & 
 \multicolumn{2}{c|}{\makebox[0pt]{with OKF}} & 
 \multicolumn{2}{c|}{\makebox[0pt]{only OKF}} & \\
\multirow{1}{*}{\makebox[0pt]{Seq\textbar obj}} & \makebox[10pt]{$E_r$(\si{\degree})} & \makebox[10pt]{$E_t$(m)} & \makebox[10pt]{$E_r$(\si{\degree})} & \makebox[10pt]{$E_t$(m)} & \makebox[10pt]{$E_r$(\si{\degree})} & \makebox[10pt]{$E_t$(m)} & \makebox[10pt]{$E_r$(\si{\degree})} & \makebox[10pt]{$E_t$(m)} \\
\midrule
\midrule
00\textbar 01 & $3.63$ & $0.38$ & $1.33$ & \textcolor{blue}{$0.26$} & \textcolor{blue}{$1.07$} & $0.35$ & $\mathbf{0.92}$ & $\mathbf{0.22}$ \\
mean & $3.72$ & $0.38$ & $1.51$ & $0.36$ & \textcolor{blue}{$1.11$} & \textcolor{blue}{$0.33$} & $\mathbf{0.79}$ & $\mathbf{0.15}$ \\
\midrule
03\textbar 01 & $3.79$ & \textcolor{blue}{$0.68$} & $3.85$ & $1.94$ & \textcolor{blue}{$0.67$} & $0.83$ & $\mathbf{0.2}$ & $\mathbf{0.15}$ \\
mean & $3.73$ & $0.72$ & $2.37$ & $1.4$ & \textcolor{blue}{$0.46$} & \textcolor{blue}{$0.46$} & $\mathbf{0.24}$ & $\mathbf{0.15}$ \\
\midrule
04\textbar 03 & $3.28$ & \textcolor{blue}{$0.9$} & $2.63$ & $1.18$ & \textcolor{blue}{$1.79$} & $1.11$ & $\mathbf{0.64}$ & $\mathbf{0.12}$ \\
04\textbar 04 & $4.13$ & \textcolor{blue}{$0.88$} & $2.1$ & $0.89$ & \textcolor{blue}{$1.39$} & $0.97$ & $\mathbf{0.38}$ & $\mathbf{0.12}$ \\
04\textbar 05 & \textcolor{blue}{$0.51$} & $1.11$ & $2.58$ & \textcolor{blue}{$1.03$} & $1.09$ & $1.09$ & $\mathbf{0.43}$ & $\mathbf{0.15}$ \\
mean & $3.74$ & $0.9$ & $2.23$ & \textcolor{blue}{$0.87$} & \textcolor{blue}{$1.92$} & $0.93$ & $\mathbf{0.76}$ & $\mathbf{0.1}$ \\
\midrule
05\textbar 20 & $2.64$ & $3.02$ & $4.58$ & \textcolor{blue}{$2.73$} & \textcolor{blue}{$2.05$} & $3.19$ & $\mathbf{0.54}$ & $\mathbf{0.23}$ \\
05\textbar 24 & $5.29$ & \textcolor{blue}{$2.74$} & \textcolor{blue}{$1.34$} & $2.79$ & $3.37$ & $3.1$ & $\mathbf{0.53}$ & $\mathbf{0.15}$ \\
mean & $2.25$ & $2.05$ & $\mathbf{0.29}$ & \textcolor{blue}{$1.94$} & $2.07$ & $2.33$ & $0.63$ & $\mathbf{0.55}$ \\
\midrule
18\textbar 04 & \textcolor{blue}{$0.78$} & $0.20$ & $1.92$ & $\mathbf{0.09}$ & $0.94$ & $0.16$ & $\mathbf{0.26}$ & $0.19$ \\
mean & $1.69$ & $1.34$ & \textcolor{blue}{$1.08$} & \textcolor{blue}{$1.00$} & $1.21$ & $1.48$ & $\mathbf{0.53}$ & $\mathbf{0.27}$\\
\midrule
20\textbar 32 & $0.77$ & $0.21$ & $0.45$ & \textcolor{blue}{$0.08$} & \textcolor{blue}{$0.35$} & $0.12$ & $\mathbf{0.08} $ & $\mathbf{0.03}$ \\
mean & $1.90$ & $0.31$ & $1.28$ & $\mathbf{0.20}$ & \textcolor{blue}{$1.33$} & $0.58$ & $\mathbf{0.68}$ & $0.53$\\
\bottomrule
\end{tabular}
\vspace{-6mm}
\end{table}

\tabref{tab:kitti_obj_pose} presents relative object pose errors which follow a similar trend to the object motion errors. 
The world-centric approach does not estimate object pose;
instead, using an initial pose,
the estimated per-frame motion can be used to propagate the pose of a corresponding object, constructing its full trajectory.
The ground truth $\prescript{\worldf}{}{\mathbf{L}}^j_{0\text{, gt}}$ is used as the starting pose.
To ensure a fair comparison, the same ground truth is used to initialise object poses $\objpose{\worldf}{0}^j$ in object-centric approaches. 
Despite not estimating for $\objpose{\worldf}{}$, 
the world-centric formulation is more accurate for \SI{\sim 95}{\percent} of all objects when compared to the state-of-the-art, 
and \SI{\sim 80}{\percent} compared to our proposed object-centric variations. 


\section{Conclusion and Future Work}
\label{sec:conclusion}

This paper has undertaken a comprehensive analysis of multiple solutions for Dynamic SLAM and evaluated the proposed formulations on existing real-world datasets. 
For that, we developed a front-end agnostic optimisation framework using GTSAM~\cite{gtsam} that can easily implement and test different configurations. 
These formulations are categorised as \emph{object-centric} and \emph{world-centric} according to how dynamic objects and their corresponding point observations are represented in the factor graph. 
The object-centric formulation is more intuitive, 
but our analysis shows that a world-centric approach produces much more accurate object motion estimations while displaying better stability during optimisation.
Our results highlight that employing different representations, as well as their subsequent graph structures, has a significant impact on the definition and performance of the underlying optimisation problem. 
In the future, we plan to derive a formal characterisation of our findings that can also be used to provide clear guidelines in advance for determining the circumstances under which specific formulations will outperform others. 



\bibliographystyle{IEEEtran}
\bibliography{./IEEEabrv, ./refs/bibliography}

\end{document}